# GRADING CONVERSATIONAL RESPONSES OF CHATBOTS

Grant Rosario[1] and David Noever[2]

PeopleTec, 4901-D Corporate Drive, Huntsville, AL, USA, 35805

[1]grant.rosario@peopletec.com [2] david.noever@peopletec.com

## *ABSTRACT*

*Chatbots have long been capable of answering basic questions and even responding to obscure prompts, but recently their improvements have been far more significant. Modern chatbots like Open AIs ChatGPT3 not only have the ability to answer basic questions but can write code and movie scripts and imitate well-known people. In this paper, we analyze ChatGPTs' responses to various questions from a dataset of queries from the popular Quora forum. We submitted sixty questions to ChatGPT and scored the answers based on three industry-standard metrics for grading machine translation: BLEU, METEOR, and ROUGE. These metrics allow us to compare the machine responses with the most upvoted human answer to the same question to assess ChatGPT's ability to submit a humanistic reply. The results showed that while the responses and translation abilities of ChatGPT are remarkable, they still fall short of what a typical human reaction would be.*

## 1. INTRODUCTION

Modern Natural Language Processing (NLP) systems have improved dramatically since the inception of the first chatbot, ELIZA, in 1966. [1] While the goal of ELIZA was to analyze human input questions and generate a human-sounding response, a plan which was surprisingly somewhat successful, it struggled with context, keyword focus, and transformations, which often made it seem robotic rather than human. The latest widely released NLP chatbot, ChatGPT, can understand the context in such a way as to offer complex code scripts [2], simulate OS terminals [3], and generate cyber security examples [4]. The feature list has grown quite a bit since the days of ELIZA; however, the purpose of this research is to focus on ChatGPT's conversational abilities.

This paper analyses the ability of ChatGPT to respond to a unique set of questions in a way that mirrors the style and speech of a human. While many methods and metrics have recently been discovered and proposed for grading an AI's natural language ability, most tend to focus on specific areas of speech, such as context [5] and response accuracy [6,7]. Our research concentrates strictly on evaluating ChatGPT responses across three metrics for grading an NLP system's conversational response: the BLEU metric [8], METEOR [9], and ROUGE scores [10]. We discuss these metrics in detail below, albeit high-level. However, we will refrain from digging too deeply into any metric and encourage the reader to look to the original algorithm papers for in-depth information. We will first provide details on the dataset used for our experiment and discuss the results.

## 2. METHODS

The experiment performed for this paper used the Quora Question Pairs dataset [11], which consisted of 60 unique questions on the Quora forum, each of which pairs with the most accepted human answer. Each question was then asked to ChatGPT via the OpenAI API using the "text-davinci-003" model representing the ChatGPT3 NLP engine released in late November 2022 [12]. Figure 1 shows the first five instances from our dataset. We then passed the responses from ChatGPT through the three metrics and used the results to assess how the response compared with the human answer, which we used as a reference.

| prompt | reference | response |
|---|---|---|
| What sport should I try out? | Table tennis is tremendously fun, and a much better source of exercise than people tend to give it credit for. It's simultaneously a very intense mental sport and one that engages a lot of different muscle groups. As an added plus, people who don't play it will think you're amazing if you can play the game at anywhere above a novice level! | That depends on what you are interested in. Some popular sports include football, basketball, soccer, baseball, tennis, golf, volleyball, swimming, and running. If you're looking for something more extreme, you could try skateboarding, snowboarding, rock climbing, surfing, or wakeboarding. |
| What is the point of a tattoo? | There are a variety of reasons for getting a tattoo. You can get one to remember someone/something, you can get one because it looks pretty or cool, or a variety of other reasons. Reasons will certainly differ from person to person. I got my tattoo to remember my grandmother, the depression I experienced after her death, and to celebrate how my partner helped me get through that dark time. | The point of a tattoo can vary from person to person. Tattoos can be used to express personal beliefs, commemorate a meaningful event or person, or simply to express style and creativity. Some people use tattoos to mark a rite of passage, while others use them to keep memories alive. |
| If we don't take revenge, how will they learn from their mistakes? | Forgive and show kindness. They will realize it themselves someday. Forgiveness is your trait. It solely depends on you and not on the other person. You want to forgive them because you want peace of mind and don't want to hold grudges against anyone in your life. Forgive because you have a strong heart. Forgive because you are strong. Strongest people are those who forgive people who aren't even sorry for hurting you. Forgive but don't forget the lessons learnt. | It is important to remember that taking revenge on someone is never the best way to respond to a mistake. Instead, it is more important to focus on teaching people the importance of learning from their mistakes and understanding why something was wrong. This can be done through open communication, education, and providing support and guidance. Ultimately, it is up to the individual to recognize and learn from their mistakes; revenge does not help them do this. |
| Do people driving Corvettes get more attention from police? | Not really. I have stopped literally thousands of cars in my career, all makes and models. We don't look for a particular model, we look and observe driving behavior. Generally, a person driving an $80,000 Corvette isn't screwing around and usually doesn't do something that would cause them to become involved in a collision. Yet, on occasion, there is an exception to every rule. | Some people may feel that they do get more attention from police when driving a Corvette, but there is no scientific evidence to support that claim. Generally, police officers are more likely to pull over vehicles that are going over the speed limit or driving recklessly, regardless of the make or model. |
| What's so scary about "being cheated on"? | Because no one wants to feel betrayed and their trust being broken. No one wants to feel stupid for being trusting enough. The most heart shattering part of cheating, which actually hurts, is "lying". When someone lies to you, you feel betrayed and that is hard to overcome."Lying is about controlling someone else's reality, hoping that what they don't know won't hurt you."Neil Strauss, The Truth: An Uncomfortable Book about RelationshipsIf your partner betrays your trust by cheating (emotionally or physically or both), that is hard to deal with because you often feel stupid, stupid enough to trust your partner in the first place. And tell you what, no one wants to feel that they are stupid in the first place. | The fear of being cheated on is rooted in a deep sense of betrayal and loss of trust. When someone cheats on their partner, it can be devastating for the victim, as it causes them to question the trust in their relationship and can even lead to feelings of worthlessness and insecurity. The fear of being cheated on can also lead to feelings of anxiety, anger, and depression. |

**Figure 1: The first five elements of the Quora Question Pairs dataset used for this experiment. Column 1: Question prompt. Column 2: Human answer, which is used as a reference. Column 3: ChatGPT response.**

### 2.1. BLEU Score

The Bilingual Evaluation Understudy (BLEU) score was initially developed to measure the performance of a machine's ability to translate text from one language to another while maintaining appropriate context and meaning [8], hence the bilingual title. BLEU can compare a machine's translation, also called the *candidate translation*, to an existing human-generated translation, known as the *reference translation*. The algorithm to compute the BLEU score works by tokenizing the candidate and calculating precision based on how many words appear in the candidate translation that also appeared in the reference translation, also called *precision*. However, this alone wouldn't be helpful since one could repeat a common word repeatedly to get a perfect score. The unique part of BLEUs precision calculation is that it begins to penalize the score when too many of the exact phrases appear in the candidate translation. The next part of the algorithm focuses on *recall*, which is the percentage of correct tokens based on the reference. The metric score penalizes candidate texts that are too short, referred to as a brevity penalty. A single BLEU score combines precision and recall, with zero being the worst and one being perfect. Typically, regarding NLP tasks, a BLEU score of at least 0.4 is considered good, while a score of 0.6 and higher is exceptional. Although its primary use was for translation, part of our proposal is that this algorithm could be beneficial for comparing output from ChatGPT to human output in a translation-like manner.

### 2.2. METEOR

METEOR, or the Metric for Evaluation for Translation with Explicit Ordering, is another translation metric but claims to have a more positive correlation with human judgment [9]. It aims to correct a weakness of BLEU that may unnecessarily penalize individual sentences due to averaged brevity. To overcome this, METEOR modifies the precision and recall by replacing them with a weighted F1-score based on mapping the unique tokens of the candidate to the reference and adding a penalty for incorrect word order. Like BLEU, the resulting METEOR score is between 0 and 1, with a higher score representing more similarity to the reference text.

### 2.3. ROUGE

Unlike the previously discussed metrics, Recall Oriented Understudy for Gisting Evaluation (ROUGE) is based solely on recall. Its typical use case is evaluating if a candidate corpus adequately summarizes a reference text [10]. The interesting characteristic of ROUGE is its different flavors for computing different types of recall. ROUGE-N is based on n-grams, so ROUGE-1 adds the recall based on matching unigrams between the candidate and reference, and this scales up based on the number of n-grams we want to compute. ROUGE-L is based on the longest common subsequence (LCS) algorithm and reports an F1 score based on the resulting precision and recall. ROUGE-W is nearly identical to ROUGE-L, the only difference being that it weighs the LCS by tracking the lengths of consecutive matches. Finally, ROUGE-S focuses on skip-bigrams, any pairs of words that allow for arbitrary gaps. Like BLEU and METEOR, the ROUGE score is a value between 0 and 1, with 0 having no similarity and 1 being a perfect match. Our experiment provides results for computing the ROUGE 1/2/L/S/W scores.

## 3. RESULTS

Figure 2 shows the results of the BLEU score calculated for each response from ChatGPT compared to the human response on the Quora forum. We can observe that with few exceptions, the AI responses do not match human responses very often.

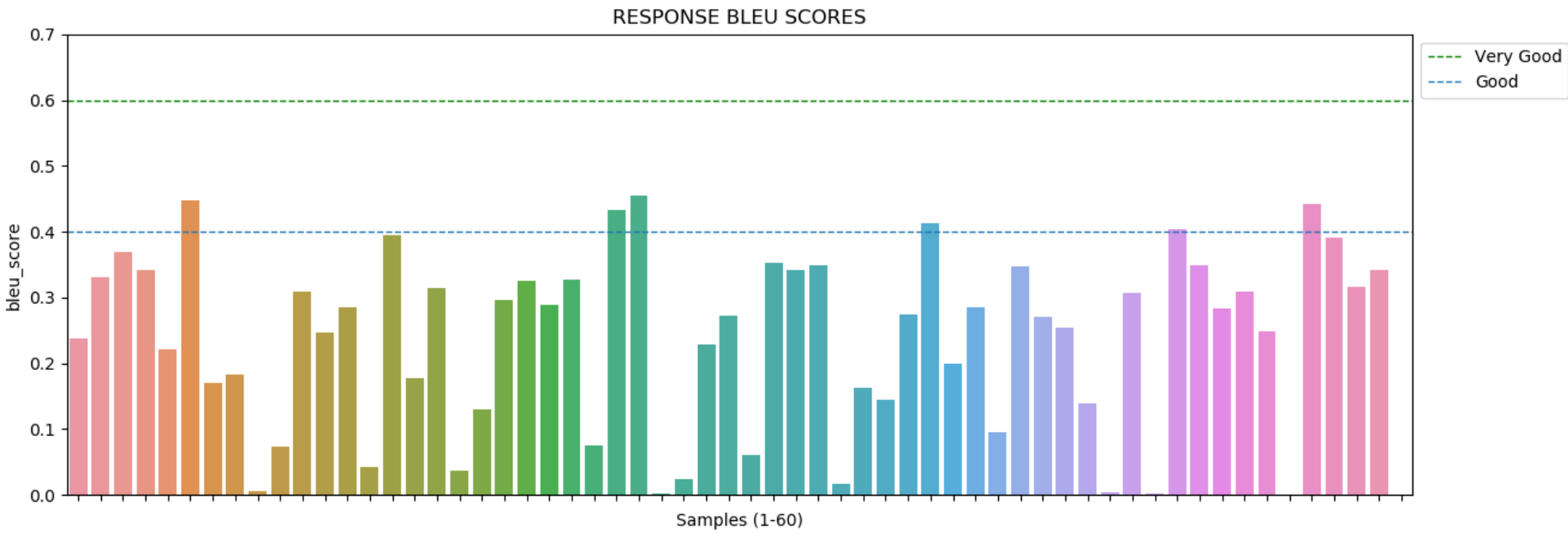

**Figure 2. BLEU score calculated for each response from ChatGPT compared to the human answer on the Quora forum**

In a similar fashion, we then computed the METEOR score for each response shown in Figure 3. Like BLEU, most answers fail to adequately match the human response well enough to be considered "human-like." However, it is interesting that the high METEOR score responses tend to correlate positively with the high BLEU score responses.

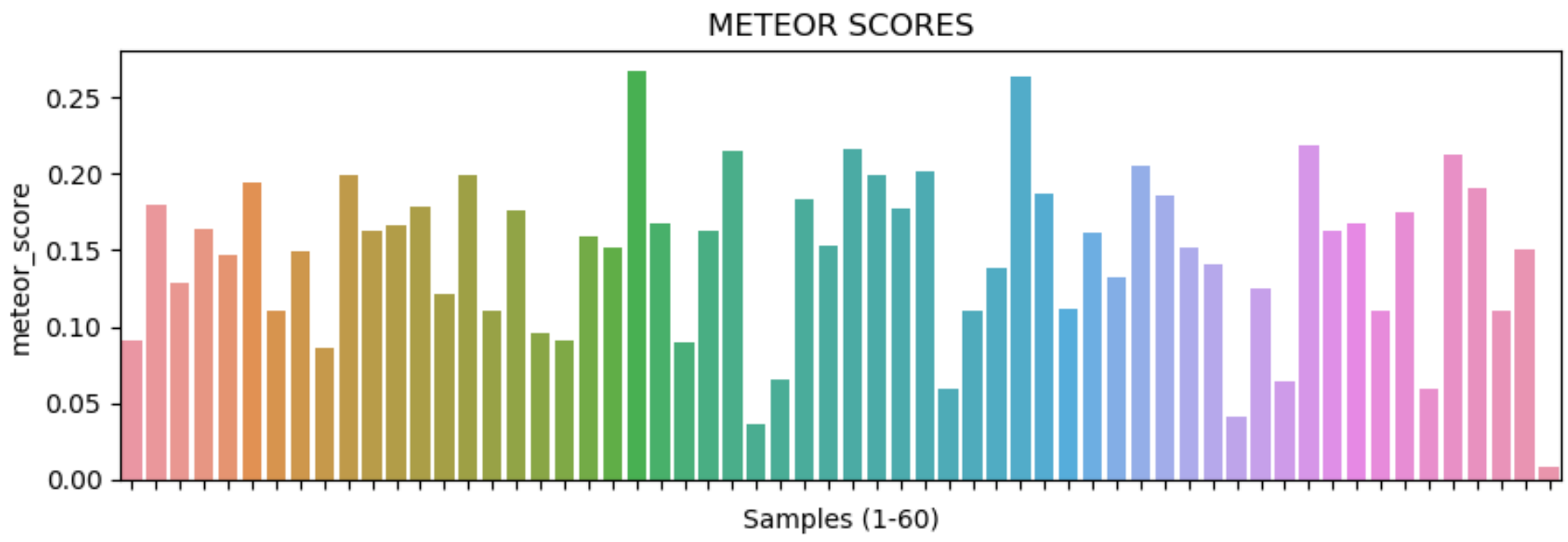

**Figure 3. METEOR score for each response**

Lastly, we show in Figure 4 an average of all the metrics we computed across all the responses. One can observe that the ROUGE scores show quite a poor similarity on average. However, we found it interesting that the average ROUGE-L outperformed the METEOR score, indicating that some of the responses must have had decent subsequence matches. However, it's important to note that this is not common.

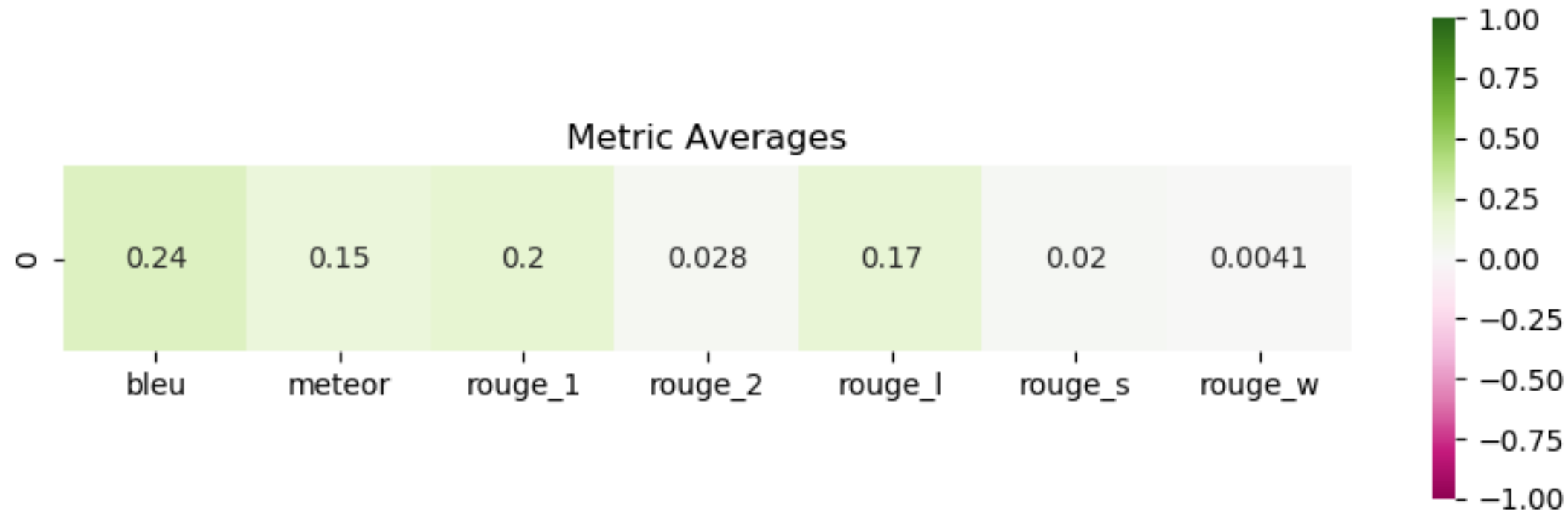

**Figure 4. Average of all the metrics we computed across all the responses**

## 4. DISCUSSION

While modern chatbots like OpenAI's new ChatGPT offer considerable functionality, this research demonstrates that its responses do not replicate the human-sounding text. Looking at our dataset, this could be because many human answers are somewhat creative in that they draw upon past experiences or other human references that an AI system has yet to accomplish well. Future work in this area could combine more metrics [13] or work to implement conversational data with a focus on question-and-answer functionality [14].

## ACKNOWLEDGMENTS

The authors benefited from the encouragement and project assistance of the PeopleTec Technical Fellows program. The authors thank the researchers at OpenAI for developing large language models and allowing public access to ChatGPT.

**Authors**

**Grant Rosario** has research experience in embedded applications and autonomous driving applications. He received his Masters from Florida Atlantic University in Computer Science and his Bachelors from Florida Gulf Coast University in Psychology.

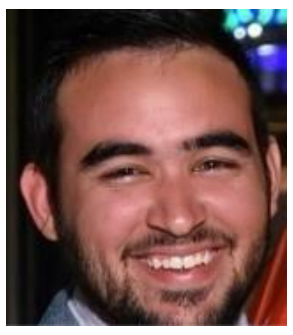

**David Noever** has research experience with NASA and the Department of Defense in machine learning and data mining. He received his BS from Princeton University and his Ph.D. from Oxford University, as a Rhodes Scholar, in theoretical physics.

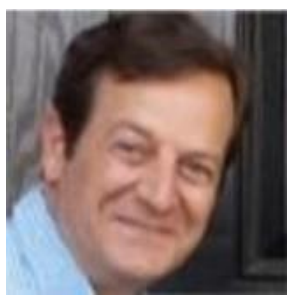